\documentclass[10pt,twocolumn,letterpaper]{article}

\usepackage{wacv}
\usepackage{booktabs} 
\usepackage{verbatim}

\usepackage{times,capt-of}
\usepackage{epsfig}
\usepackage{graphicx}

\usepackage{amsmath}
\usepackage{amssymb}
\usepackage{balance}
\usepackage{lipsum,adjustbox}

\usepackage{hyperref}

\wacvfinalcopy 

\def\wacvPaperID{343} 

\ifwacvfinal\fi
\begin{document}

\title{ Supervised and Unsupervised Learning of Parameterized Color Enhancement}


\author{Yoav Chai\\
Computer Science\\ Tel Aviv University
\and
Raja Giryes\\
School of Electrical Engineering\\ Tel-Aviv University
\and
Lior Wolf\\
Facebook AI Research and\\
Computer Science, Tel Aviv University}
\twocolumn[{%
\renewcommand\twocolumn[1][]{#1}%
\maketitle
\begin{center}
  \centering
  \vspace{-.8cm}
  \includegraphics[width=\textwidth]{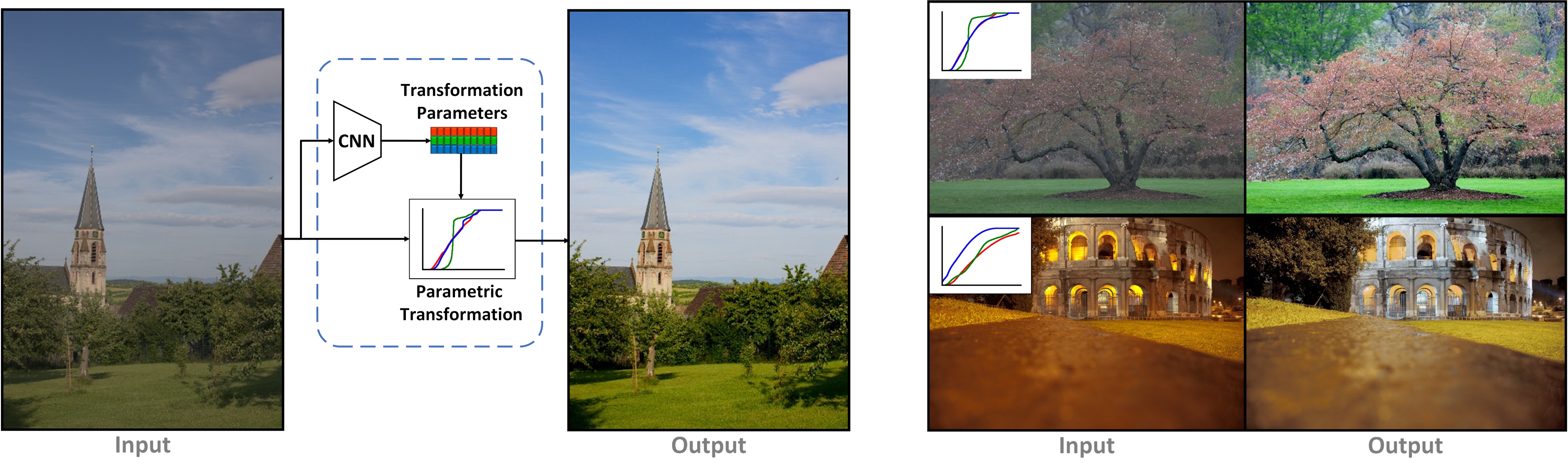}
  \captionof{figure}{
  Our method provides automatic image retouching. It is a learning-based technique that can be trained using either paired or unpaired images. 
Once learned, the input image is fed to a CNN that determines the coefficients of a parametric color transformation that is applied to the input image. The channel curves are shown in a simplified way as a mapping of R,G,B values.
  }
  \label{fig:teaser}

\end{center}%
}]
\ifwacvfinal\thispagestyle{empty}\fi

\begin{abstract}
We treat the problem of color enhancement as an image translation task, which we tackle using both supervised and unsupervised learning. Unlike traditional image to image generators, our translation is performed using a global parameterized color transformation instead of learning to directly map image information. 
In the supervised case, every training image is paired with a desired target image and a convolutional neural network (CNN) learns from the expert retouched images the parameters of the transformation. In the unpaired case, we employ two-way generative adversarial networks (GANs) to learn these parameters and apply a circularity constraint. We achieve state-of-the-art results compared to both supervised (paired data) and unsupervised (unpaired  data) image enhancement methods on the MIT-Adobe FiveK benchmark. Moreover, we show the generalization capability of our method, by applying it on photos from the early 20th century and to dark video frames.
\end{abstract}

\section{Introduction}

The number of captured photos has grown steadily since the advent of phone cameras. In many cases, casual photos require additional editing in order to enhance their quality. While photo editing programs provide various retouching operations, they require expertise. In addition, the manual editing process may become time-consuming, depending on the initial quality.  


In this work, we focus on the task of color enhancement. Color enhancement of raw images significantly improves the quality for an observer~\cite{bianco2009image}. Yet, an automatic color enhancement is a non-trivial task because it depends on content, context and color distribution.

Color enhancement transformations can be performed by a global parameterized transformation or by local modifications. A global transformation applies the same operation to the whole image and usually uses a smooth function to preserve the quality of the image. Its advantage is the support of an arbitrary image resolution in a coherent manner. Local operations depend on the local content of the image and often use highly non-linear mappings that lead to artifacts.


{\bf Contributions.} 
In this work, we pursue an image enhancement transformation which enjoys the following desired properties: (i) preservation of the quality of the image;  (ii) possibility to be applied to an arbitrary resolution and specifically, high-resolution images; (iii) consumption of minimal computational resources; (iv) flexibility in training: the method can be applied with and without paired data samples. While some of these characteristics are present in the existing solutions, we are not aware of any work that incorporates them all together. Our method obtains state-of-the-art results on the existing image enhancement benchmarks in both the supervised and unsupervised cases. Particularly, we show through a user study that our strategy outperforms the leading techniques in the MIT-Adobe FiveK raw-images enhancement task and a commercial software on the enhancement of old color photos.

In the case of supervised learning, one benefits from the existence of a pair of an input and a retouched image. In the unsupervised case, training is based on two unmatched sets: a set of input images and a set of retouched (high quality) images. Unsupervised learning has the advantage that the two sets can be collected independently. For example, the input images can be taken from collections of casual photography, while high quality images can be collected from stock photo websites.


The color transformation we use is parameterized as a quadratic mapping of the color information, i.e., as a linear mapping of the set of all second order monomials of the RGB values. The coefficients of this mapping are estimated with a CNN. One can thus view our CNN as a function that maps images to quadratic functions. In the supervised case, this function is optimized with a direct loss. In the unsupervised case, it learns the mapping in both directions, and employs a circularity constraint. To improve the training in the latter case, we introduce to it several improvements, such as weight sharing between the two generators but without sharing the Batch Normalization~\cite{ioffe2015batch} parameters, and a novel multi-phase training technique. 

Unlike previous GAN-based image-to-image enhancement techniques, e.g.,  \cite{Chen:2018:DPE}, our generator can enhance images for an arbitrary resolution without suffering from artifacts that appear in previous GAN based solutions. This is achieved because our network determine the parameters of a global adjustment operation from a low-resolution version of the image and apply it on the full-resolution image, instead of being applied directly as an image to image mapping.
Compared to reinforcement learning methods that have been used for image enhancement~\cite{park2018distort,hu2018exposure}, and require a sequential application of a deep network, we need just a single pass of a relatively simple CNN. The gain in efficiency that arises from these advantages makes our method especially suitable for running on limited resource devices, such as low-end smart-phones and on the camera itself.


\section{Related Work}

Several research works considered the problem of automatic color enhancement. They may be divided into two types: (i) Example-based methods transferring the color of an example image to a given input image; and (ii) Learning-based techniques that use a training data to find a mapping function from the input image to a target image. Some of these works can be trained only in the presence of paired data, which limits their usage to the supervised case.


Among the example-based methods, Reinhard \etal~\cite{reinhard2001color} and Faridu \etal~\cite{faridul2014survey} present a method where the global color distribution of an input image is warped to mimic an example style. Some recent works use image retrieval methods to achieve a (semi-)automate exemplar selection, which improves the matching \cite{hwang2012context},\cite{lee2016automatic}, \cite{liu2014autostyle}. While the example-based strategy can provide expressive enhancement and diverse stylizations, the results greatly depend on having proper example images. Moreover, even if they are given, the matching of input images to good example images is a challenging task.  

We focus here mainly on learning based methods, which lead to improved results \cite{Chen:2018:DPE,park2018distort,yan2016automatic,gharbi2017deep,bianco2019learning,hu2018exposure}. Many of these contributions, especially those applying supervised learning, employ the MIT-adobe FiveK dataset presented by Bychkovsky \etal~\cite{bychkovsky2011learning}. It contains retouched pairs of images which are created by five professional experts.


Among the learning-based color enhancement methods, there are learned local transformations \cite{yan2016automatic, gharbi2017deep, Chen:2018:DPE}. Yan \etal~\cite{yan2016automatic} have proposed a model that given the color, global and local features of an image, such as semantic content and object detection, a deep neural network maps each pixel to achieve the desired style. Gharbi \etal~\cite{gharbi2017deep} have trained a model to predict local affine transforms in the edge/color bilateral space, which can serve as an approximation to edge-aware image filters and color/tone adjustments for real-time image enhancement. 

In the more general field of cross-domain image translation, Isola \etal~\cite{isola2017image} proposed a conditional adversarial network as a general-purpose solution to image-to-image translation problems, converting from one representation of a scene to another, e.g., from a semantic label map to a realistic image or from a day image to its night counterpart.
Although generating very good results, their method requires paired images for training. Two-ways GANs were later proposed and introduced cycle consistency to address this problem. Examples of such GAN based solutions include CycleGAN \cite{CycleGAN2017}, DiscoGAN \cite{kim2017learning} and DualGAN \cite{yi2017dualgan}.


Based on these strategies, Chen \etal~\cite{Chen:2018:DPE} train a one-way and two way "traditional" GAN for image to image enhancement with $512\times512$ resolution images. 
However, it fails to perform enhancement on arbitrary resolutions, since an image enhancement operation needs to be determined by examining the content and context of the entire image. While their generator is fully convolutional, since it has a limited receptive field it does not capture the whole image when it is of high resolution. Another drawback of such a local color operation is the possible artifacts and relatively high memory consumption it has when applied to a high resolution image.


As an alternative, global parameterized transformations have been recently proposed for color enhancement \cite{park2018distort,hu2018exposure, bianco2019learning} and image processing \cite{schwartz2018deepisp}. Hu \etal~\cite{hu2018exposure} learn a one-way GAN based agent, which is trained using reinforcement learning (RL) with an adversarial reward to perform a sequence of image enhancement operations such as modifying the contrast or the brightness.

Park \etal~\cite{park2018distort} train an agent using RL to enhance the input image using similar image enhancement operations in both the supervised and unsupervised cases. For unpaired learning, they propose the {\em distort and recover} training scheme, which generates an input image for training by distorting retouched images. The model is trained to map the distorted images back to the originally retouched ones. Although achieving improved results, given an unpaired training set, the scheme requires adjustment of the distortion procedure such that it produces distorted images with similar characteristics as the input images in the set.


Bianco \etal~\cite{bianco2019learning} recently proposed a different enhancement method, which generates a series of parametric operations in color space such as polynomial, piecewise linear, cosine, and radial functions, where the CNN network determines how they are applied in order to perform the desired color transformations.


\section{Method}
 
Our photo enhancer model $G_X$ takes an input image $x$ and generates an output image $G_X(x)$ as the enhanced version of it.  
The model is a per-pixel color transformation designed to preserve the quality of the image and requires limited computation at full resolution.
To enhance the color of an input image, a fixed sized down-sample version of an input image $x$ is forwarded to a CNN $H$ to infer the parameter matrix $\theta_x$ of the color transformation. The color transformation is 
applied independently to each pixel in $x$.


 
\subsection{Parametrized Quadratic Transformer}

The image enhancement method we propose is based on a CNN, defined as $H$, which determines the coefficients of a global image enhancement operation. We formulate $H$ as a parametric function $\theta_x = H(\theta,x)$, where $\theta$ represents the CNN parameters, $x$ the input image, and $\theta_x$ the parameters of the transformation of the color space for image $x$.

To get a transformation that preserves the image high frequencies and content without artifacts but yet is powerful enough to mimic the image enhancement transformation produced by image editing software, we use color basis vectors. Let $V(p)$ be the color basis vector for a pixel with RGB values $p=[R,G,B]$. In our work, we define $V(p)$ to be a $1\times 10$ quadratic color basis vector $[R,G,B,R^2,G^2,B^2,R\cdot G,G \cdot B,B\cdot R,1]$, where R,G,B are the corresponding RGB values at pixel $p$.

For an input image $x$, $H$ outputs a matrix $\theta_x\in \mathbb{R}^{10\times 3}$, which contains the coefficients of the quadratic transformation applied to each pixel in $x$. We perform an additive correction, and learn the residual transformation, which is known to speed up training, compared to learning the target function from scratch.
Thus, the RGB value in the output image $\bar y_k$ of same location as pixel $p$ in the input image $x_k$ is:
\vspace{-.3cm}
\begin{equation}
\begin{split}
\bar p = & H(\theta,x_k) V(p) + p.
\end{split}
\end{equation}

In order to be invariant to the resolution of the input image, we use a $256\times256$ scaled (typically down-sampled) version of the images as the input of the CNN, $H$. The color transformation is then applied to all image pixels of the original image, regardless of its resolution. The ability to perform the transformation at every image resolution is a clear advantage in comparison to other literature approaches, some of which rely on an encoder-decoder architecture with a fixed output resolution.

\subsection{Training for paired data}
\label{sec:training_paired_data}

For paired learning, we used a five-branch processing network. Each branch consists of a stack of convolutional layers that ends by average pooling as a feature extractor and two linear layers. We further process the output of the five branches with another linear layer. Due to the relatively small size of the paired training set, we add Dropout~\cite{srivastava2014dropout} to the network to prevent over-fitting. A figure presents the architecture of our generator for the paired model attached in supplementary material


Figure~\ref{fig:supervised} depicts our paired training scheme. We optimize the output of the transformer with the mean $L_2$-loss CIELab color space between the transformed image and the expert retouched image.
$L_2$ in CIELab color space is defined as the Euclidean distance:
\begin{equation}
\label{eq:L2}
 L_2 = \sqrt{(l_1-l_2)^2 + (a_1-a_2)^2 + (b_1+b_2)^2},
\end{equation}
where $(l_1; a_1; b_1)$ and $(l_2; a_2; b_2)$ are the coordinates of a pixel in the two images
after their conversion from RGB to CIELab. In our case, the two images are $\bar y_k=G(x_k)$ and $y_k$.

\begin{figure}[t]
    \center
  \includegraphics[width=0.5\textwidth]{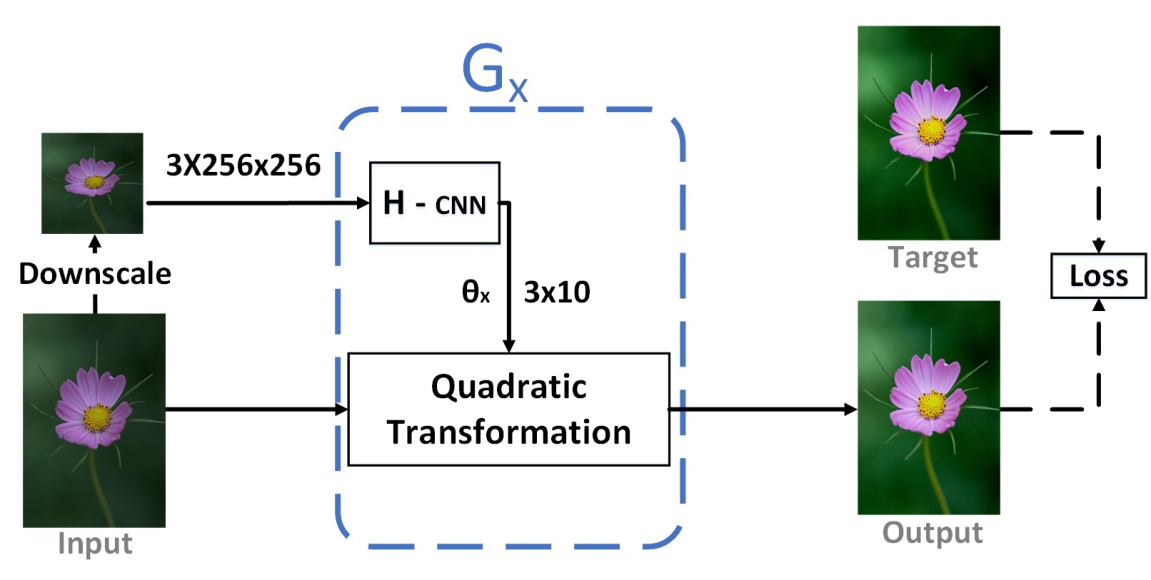}
  \caption{ Pipeline of the proposed method for paired training. The input image is down-sampled and fed to a CNN that determines the coefficients for our quadratic parameterized color transformation to be applied to the input image. At test time, the quadratic transformation is applied to the full resolution image.
  }
  \label{fig:supervised}
\end{figure}

\subsection{Training for unpaired data}
For unpaired data, the problem can be naturally formulated using the framework of two-way GANs, which learn the embedding of the input samples and generate output samples located within the distribution of the target samples. Such frameworks have been frequently used to address the image-to-image translation problem, which transforms an input image from the source domain $X$ to an output image in the target domain $Y$. Each GAN model often consists of a discriminator $D$ and a generator $G$. In our case, $G$ has a specific parametric form.

Recall that the source domain $X$ contains the input images, while the target domain $Y$ contains unmatched images with the desired characteristics. To match between the two in the absence of direct pairing, two-way GANs such as CycleGAN \cite{CycleGAN2017} and DualGAN \cite{yi2017dualgan} enforce cycle consistency, which creates a pairing between the sets $X$ and $Y$. Two-way GANs often contain a forward mapping $G_X: X \rightarrow Y$ and a backward mapping $G_Y: Y \rightarrow X$. The cycle consistency conditions require that $G_Y(G_X(x)) = x$ and $G_X(G_Y(y)) = y$, where the generator $G_Y$ takes a $G_X$-generated sample and maps it back to the source domain $X$, while the generator $G_X$ takes a $G_Y$-generated sample and maps it back to the target domain $Y$. 

\begin{figure}
    \center
  \includegraphics[width=0.5\textwidth]{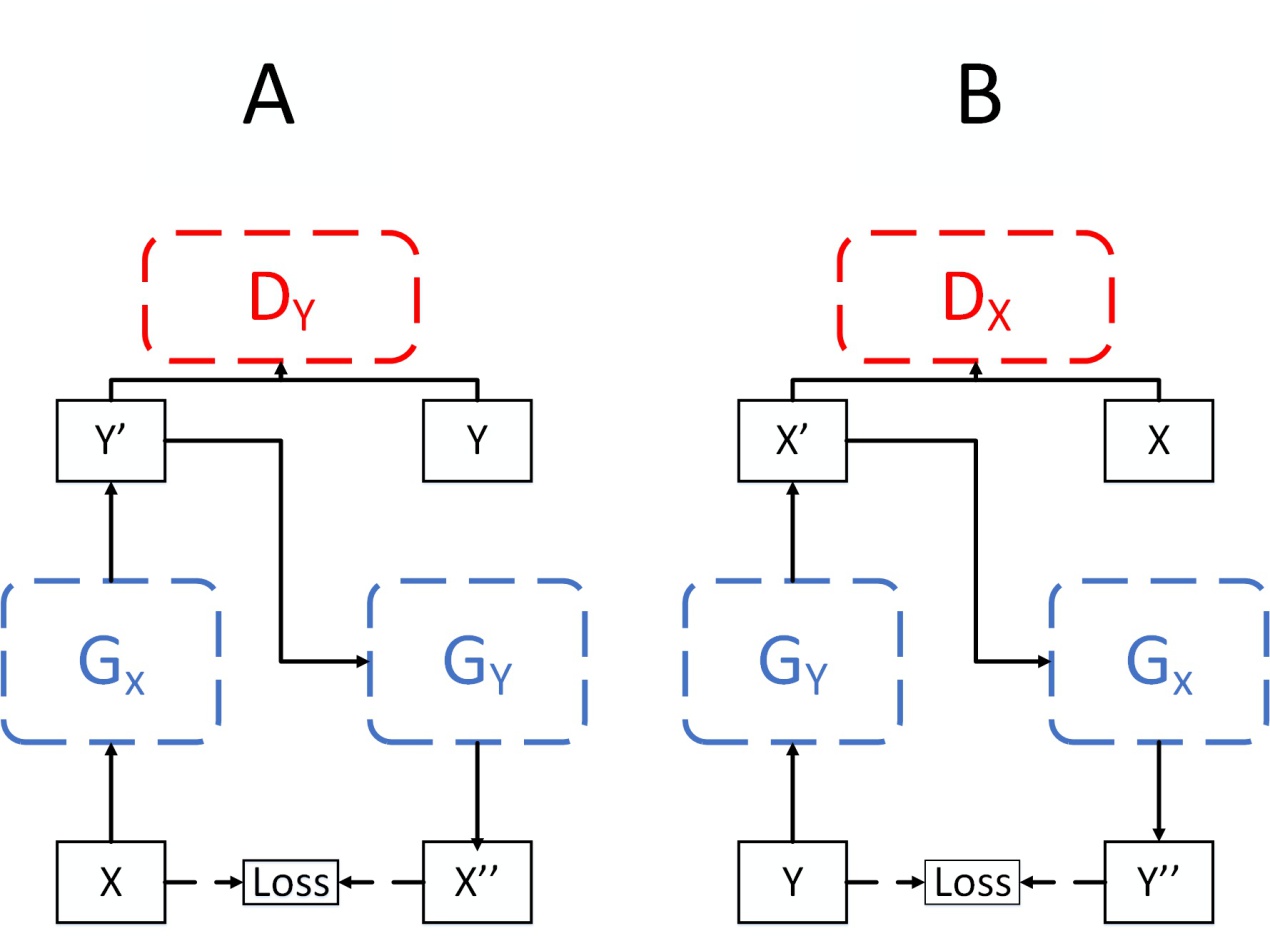}
  \caption{The architecture of our two-way GANs.}
  \label{fig:unsupervised}
\end{figure}

Figure~\ref{fig:unsupervised} describes the architecture of a two-way GAN. The discriminator $D_Y$ aims at distinguishing between the samples in the target domain ${Y}$ and the generated samples ${y'=G_X(x)}$, while the discriminator $D_X$ aims at distinguishing between the samples in the target domain ${X}$ and the generated samples ${x'=G_Y(y)}$.
We define by $x''=G_Y(G_X(x))$ the output for a  training image $x \in X$, which passes through both generators, and in a similar way we define $y''=G_X(G_Y(y))$  for an image $y \in Y$.

  

The cycle consistency losses are given by:
\begin{equation}
\begin{split}
\mathcal{L}_{cycleX} = ||x^{''} - x ||_2 \\
\mathcal{L}_{cycleY} = ||y^{''} - y ||_2,
 \end{split}
\end{equation}
where $||\cdot||_2$ is the Euclidean $L_2$ norm (see Eq.~\eqref{eq:L2}). 

The adversarial loss is given by
\begin{equation}
\begin{split}
\mathcal{L}_{GAN}(G_x,D_Y,X,Y)=&E_{y \sim p_{data(y)}} ) [\log D_Y (y)] +\\ & E_{x \sim p_{data(x)} } [1-\log D_Y (G_x (x))].
\end{split}
\end{equation}
We did not employ the identity term used in many image to image translation frameworks, due to the different distribution characteristics of the datasets $X$ and $Y$ and the usage of Batch Normalization in our networks.

To summarize, the loss we use for the unpaired training is
\begin{eqnarray}
&& \alpha \mathcal{L}_{cycleX}+ \alpha \mathcal{L}_{cycleY}+ \\ \nonumber &&
L_{GAN} (G_x,D_Y,X,Y)+L_{GAN} (G_y,D_X,Y,X),
\end{eqnarray}
where $\alpha$ is a scaling parameter. 

The generator architecture for unpaired learning is similar to the one used for paired learning but is deeper and with only one branch to reduce the overall number of parameters in it. Moreover, in the training of the network, we do not use the Dropout layer at the initial training (we add another training phase described in Section~\ref{sec:improving_unpaired_training}, where we do apply the Dropout).

Our discriminator architecture is the same one used in CycleGAN~\cite{CycleGAN2017}. It consists of a stack of convolutional layers followed by Batch Normalization, a Leaky rectified linear unit (ReLU) \cite{maas2013rectifier} and Dropout. A figure describes both the generator and the discriminator architectures used in our unpaired model attached in supplementary material.

\subsection{Improving unpaired training}
\label{sec:improving_unpaired_training}

\begin{figure}
    \center
  \includegraphics[width=0.37\textwidth]{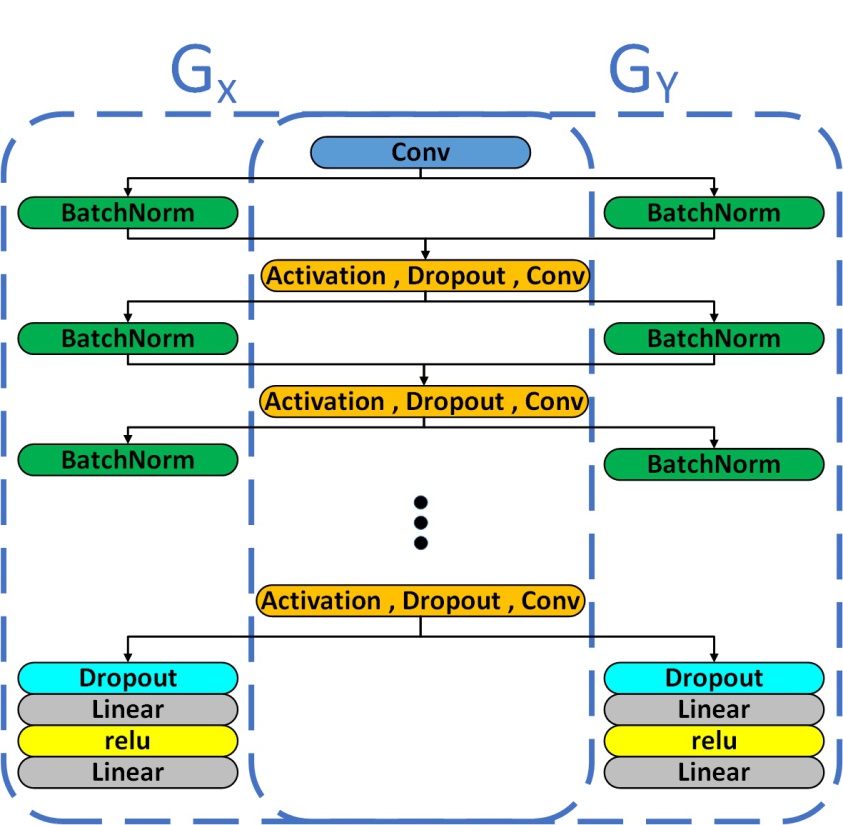}
  \caption{Illustration of the shared weights in the generators $G_X$ (left) and $G_Y$ (rigth) in the unpaired case. Notice that the parameters of Batch Normalization and the linear layers are not shared.}
  \label{fig:shared_paramaters}
\end{figure}

In order to improve the result of the unsupervised method, we use the following modifications. First, we share the feature extractor weights of the generators $G_X$ and $G_Y$, except of the linear layers and the Batch Normalization parameters, which should be different due to the different distribution characteristics of the datasets $X$ and $Y$.  Figure~\ref{fig:shared_paramaters} describes the transformers' shared parameters.

In addition, we introduce a novel cycle-consistency training, which we refer to as the second phase. In this phase, we apply the Dropout in $G_X$ and $G_Y$ (the location of the Dropout layers described in the architecture figure in supplementary material), 
but do not share their weights and use only the cycle-consistency losses. 
The motivation is to increase the generalization of the generator to random perturbation generated by Dropout.
Due to using the cycle-consistency losses exclusively (without the regular GAN loss), we train only $G_X$ and freeze $G_Y$. Training both generators simultaneously, without the GAN losses, will end up by simply outputting the input image as this leads to a zero loss. 
Thus, when training $G_X$ we use only the loss term $ || y^{''} - y ||_2$ (see part A in Figure~\ref{fig:unsupervised}).

This training technique can be seen as an improvement of the {\em distort and recover} scheme proposed by \cite{park2018distort}, but unlike them we learn the ``distortion'' (mapping from retouched images to the raw images) and ''recovery'' (mapping from the raw to the retouched) using the first phase and our second phase optimizes the generator to perform these tasks with some random perturbations.






\begin{figure*}

\newcommand{\tfig}{1.9}

\setlength\tabcolsep{1pt}

\begin{tabular}{c c c c c c c }

\includegraphics[width=0.1428\textwidth]{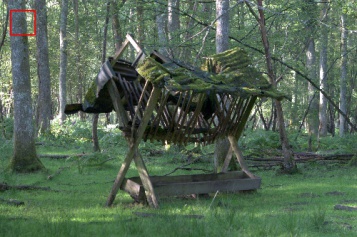} &
\includegraphics[width=0.1428\textwidth]{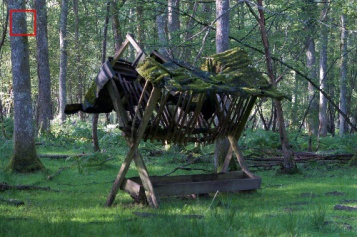}&
\includegraphics[width=0.1428\textwidth]{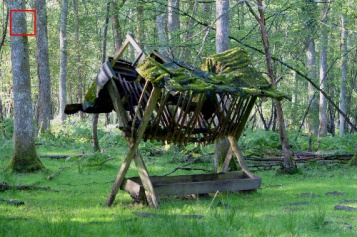} &
\includegraphics[width=0.1428\textwidth]{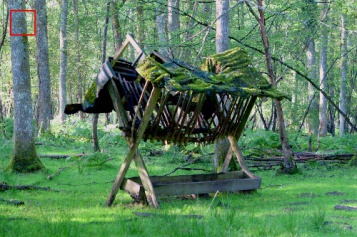}&
\includegraphics[width=0.1428\textwidth]{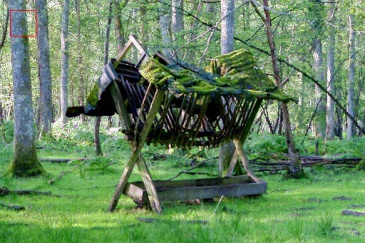}&
\includegraphics[width=0.1428\textwidth]{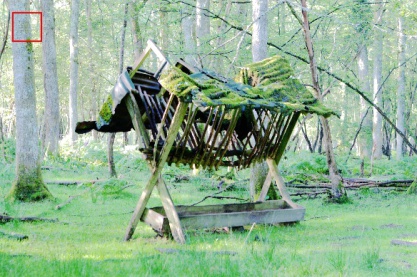} &
\includegraphics[width=0.1428\textwidth]{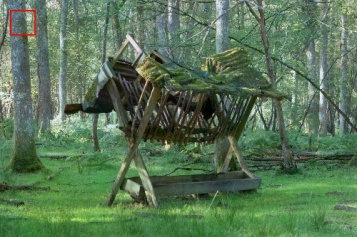} 
\\

\includegraphics[width=0.1428\textwidth]{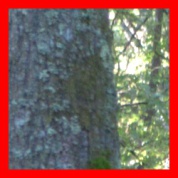} &
\includegraphics[width=0.1428\textwidth]{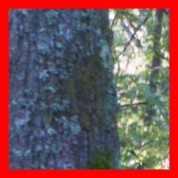}&
\includegraphics[width=0.1428\textwidth]{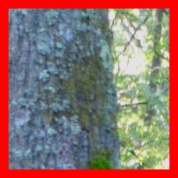} &
\includegraphics[width=0.1428\textwidth]{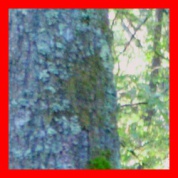}&
\includegraphics[width=0.1428\textwidth]{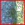}&
\includegraphics[width=0.1428\textwidth]{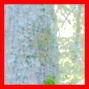} &
\includegraphics[width=0.1428\textwidth]{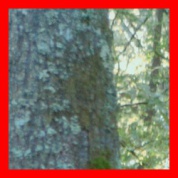} 
\\


\includegraphics[width=0.1428\textwidth]{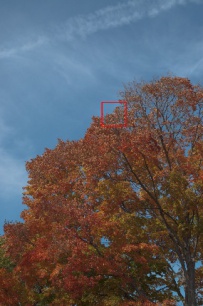}&
\includegraphics[width=0.1428\textwidth]{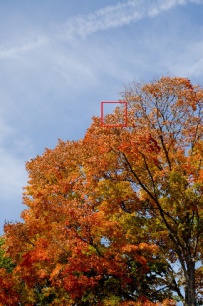} &
\includegraphics[width=0.1428\textwidth]{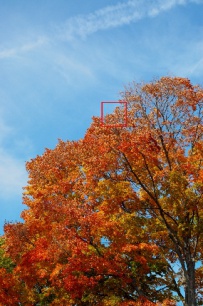} &
\includegraphics[width=0.1428\textwidth]{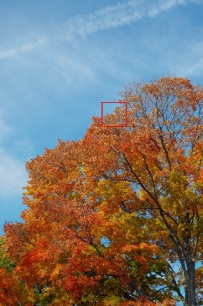}&
\includegraphics[width=0.1428\textwidth]{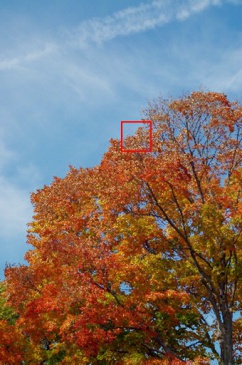}&
\includegraphics[width=0.1428\textwidth]{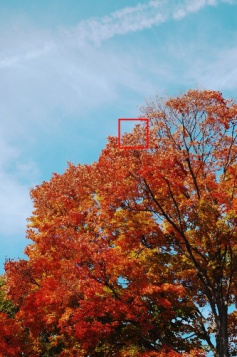}&
\includegraphics[width=0.1428\textwidth]{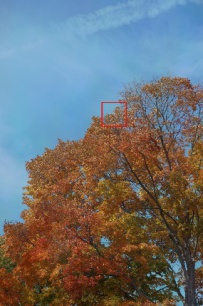} 
\\
\includegraphics[width=0.1428\textwidth]{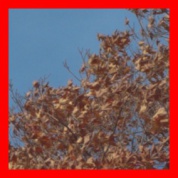}&
\includegraphics[width=0.1428\textwidth]{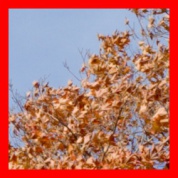}&
\includegraphics[width=0.1428\textwidth]{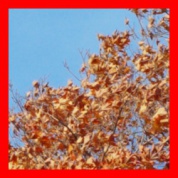} &
\includegraphics[width=0.1428\textwidth]{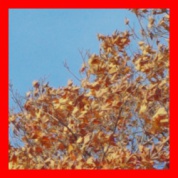}&
\includegraphics[width=0.1428\textwidth]{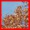}&
\includegraphics[width=0.1428\textwidth]{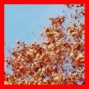} &
\includegraphics[width=0.1428\textwidth]{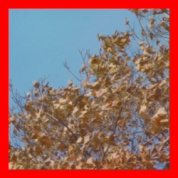} 

\\

input &
expert &
ours &
ours  &
DPE &
Exposure &
HDR-net

\\
 &
 &
 paired &
unpaired &
 unpaired &
 unpaired &
 paired
 \\
 &
 &
  &
  &
\cite{Chen:2018:DPE} &
\cite{hu2018exposure} &
\cite{gharbi2017deep}

\\
\end{tabular}    

\caption{Qualitative comparison with \cite{Chen:2018:DPE,hu2018exposure,gharbi2017deep}. We compare both full image and the zoom-in version of the image.}

\label{fig:comprassion_zoom}
\vspace{-3mm}

\end{figure*}


\section{Experiment}

We evaluate our model's ability to reproduce human-annotated retouches through extensive experiments:
    (1) Evaluation on
input-retouched paired dataset in Section.~\ref{sec:exp_paired_data}.
(2) Evaluation on input-retouched unpaired dataset in Section~\ref{sec:exp_unpaired_data}.
(3)  Ablation study of our proposed training techniques in Section~\ref{sec:exp_ablation_study_paired}
and Section~\ref{sec:exp_ablation_study}.
(4) Evaluation of the paired and unpaired method with a user study in Section~\ref{sec:user_study}.


To assess the accuracy of our proposed method, we used the MIT-Adobe FiveK dataset, which contains 5,000 high resolution raw images in DNG format, and their corresponding manually retouched images by five variants; each retouched by a different expert A/B/C/D/E. Expert C got the highest Mean Opinion Score(MOS), so we follow the common practice and use the images created by this expert as the target. 

We use the procedure described by \cite{hu2018exposure} to preprocess the raw images and export them to the SRGB format \cite{susstrunk1999standard}. As a performance measure, we calculate the mean $L_2$ on the CIELab color space \cite{fairchild2013color}, and the SSIM \cite{wang2004image} and PSNR in the RGB color space. All the errors are with respect to the results of expert C on the test set images.

\subsection{Paired data}
\label{sec:exp_paired_data}

We split the MIT-adobe FiveK dataset into train and test images. Each literature method splits the data differently and the train-test split of some leading methods is not available. For a fair comparison, we chose a train-test split that contains the smallest amount of training images in the literature, which is 4000. To reduce the randomness that arises from the splitting process, we randomly split the data five times, trained our model and averaged the measured performance.
We also show our result on the exact same train-test split used by Chen \etal~\cite{Chen:2018:DPE}.

We trained the model with a batch size of 50 and the Adam optimizer \cite{kingma2014adam} for 500 epochs. We used a base learning rate $9 \cdot 10 ^{-4}$ and linearly decayed it every 30 epochs up to $2 \cdot 10 ^{-6}$ at epoch 300. We used rotations and vertical flips as augmentation.

Table~\ref{tab:exp_supervised} compares our method with all leading paired training methods. Since not all experiments in the literature are done under the same conditions, we detail in the table, the transformation method, train-test split, and the image resolution used for evaluation, in addition to the mean $L_2$ in CIELab color space, SSIM and PSNR in RGB color space. Our method achieves better prediction performance in terms of mean $L_2$, SSIM and PSNR than all the other methods. When comparing to global transformation techniques, our margin is even higher. When comparing to local approaches, the $L_2$ measure does not capture the local artifacts created by these methods, which is a known limitation they have \cite{yan2016automatic,gharbi2017deep,Chen:2018:DPE}.

\begin{table*}
		\centering
			\begin{center}
					\begin{tabular}{lccccccc}
						\toprule
						
                        
                        method & $\downarrow$mean $L_2$ & $\uparrow$SSIM & $\uparrow$PSNR & color  & Train-Test split & Resolution  & number of\\
                        & & & & operation &   &   &  forward path\\
                        
				 		\midrule
				 		
						{Ours} & \bf{9.40}& \bf{0.920} & \bf{23.93}& global & 4000-1000 & Full-Size & 1\\
												
						{Learning Parametric  } & 10.36& NA & NA & global & 4000-1000 & Full-Size & 1\\
 					 {Functions \cite{bianco2019learning}} & &  &  &  \\
						
 					{Distort-and-Recover \cite{park2018distort}} &10.99& 0.905 & NA & global & 4750-250 & NA & multiple\\

 					 {Automatic Photo \cite{yan2016automatic}} & 9.85& NA & NA & local & 4750-250 & Full-Size & per pixel\\

 					 {HDR-Net \cite{gharbi2017deep}} & 12.14& NA & NA &local & 4750-250 & Full-Size & 1\\
 					 \midrule
 					  					
 					  					 					  					{Ours, splits of~\cite{Chen:2018:DPE}} & \bf{9.05} & \bf{0.937} & \bf{24.11} & global & 4500-500 & 512 long side & 1\\
 					
 					{Photo Enhancer \cite{Chen:2018:DPE}} & NA& NA & 23.8& local & 4500-500 & 512 long side & 1\\
						\bottomrule
					\end{tabular}

					\caption{
					Comparison of leading paired base methods evaluated by the mean $L_2$, SSIM and PSNR with respect to the expert target using the MIT-Adobe FiveK dataset.  }
					\label{tab:exp_supervised}
			\end{center}
			\vspace{-4mm}
\end{table*}


\subsubsection{Alternative architectures}
\label{sec:exp_ablation_study_paired}
In order to evaluate our multi-branch processing architecture for paired learning as detailed in Sec.~\ref{sec:training_paired_data}, we report the performance of the following setups. {\em Complete Method:} our proposed model; {\em Three branches generator:} our model but with three branches; {\em Single branch generator } our best single branch architecture for paired learning, which we also use for unpaired learning.

We evaluated the result on a train-test split 4000-1000 using $256\times256$ resolution image  from the MIT-Adobe FiveK dataset. Table~\ref{tab:ablation_paired} demonstrates the effect of the multi-branch processing.

\begin{table}
		\centering
					\begin{tabular}{lc}
						\toprule
                        
                        method & $\downarrow$mean $L_2$ \\
                        
				 		\midrule
				 		
						{Complete Method} & 9.05 \\
												{Three branches generator} & 9.15 \\
						
						Single branch generator & 9.40\\
 
						\bottomrule
					\end{tabular}
					\caption{
				Evaluation of our multi-branch processing for paired learning. We report the mean $L_2$ for the three architectures on the 1000 testing images from the MIT-Adobe FiveK dataset using $256\times256$ image resolution. Note that the results of Tab.~\ref{tab:exp_supervised} are given for the full resolution.}
                \label{tab:ablation_paired}
\end{table}





\subsection{Unpaired Data}
\label{sec:exp_unpaired_data}

We split the MIT-adobe FiveK dataset into $n$ training images and $5000-n$ test images. The training images split into $n/2$ unpaired raw images and $n/2$ different target images. Thus, we make sure that the training dataset does not contain a pair of both raw and the corresponding retouched image. Without this separation, an unsupervised method may implicitly match the images and then rely on a form of supervised training. 
For a fair comparison, we use the exact same train-test split as in \cite{park2018distort,Chen:2018:DPE}, which are the current leading methods for global and local transformation for unpaired learning, respectively. We also use a random split of the same size reported by~\cite{hu2018exposure}, in the absence to the exact splits.

The first phase was trained with batch size of 20 with the Adam optimizer for 200 epochs. We kept the learning rate at $10^{-4}$ for the first 100 epochs and linearly decayed the rate to zero over the next 100 epochs. We used rotations and vertical flips for augmentation. The discriminator Dropout value was set to 0.12 and $\alpha$ was set to 0.02.

For the second phase, we trained using batch size of 50. We kept the learning rate at $5 \cdot 10^{-6}$ for the first 100 epochs and linearly decayed the rate to zero over the next 100 epochs. The generators Dropout value was set to 0.15, while other hyper-parameters remained unchanged.

Table~\ref{tab:exp_unsupervised} compares our method with all leading image enhancement methods that are trained with unpaired data. Comparison is done with mean $L_2$ in LAB space, SSIM and PSNR in RGB space. In addition, we detail the transformation method, train-test split, and the image resolution used for evaluation. Our method achieves better prediction performance in terms of $L_2$, SSIM, and PSNR, and presents our flexibility for arbitrary resolutions, and specifically, high resolution outputs.

Figures~\ref{fig:comprassion_zoom} 
show examples of our method, in both the paired and unpaired training scenarios, compared with expert C and the leading  enhancement methods available, more results attached as supplementary material.



\begin{table*}[t]
		\centering
			\begin{center}
					\begin{tabular}{lccccccc}
						\toprule
                        
                        method & $\downarrow$mean $L_2$  & $\uparrow$SSIM & $\uparrow$PSNR &  color  & Train-Test split & Resolution & number of \\
                        
                          &    &  &  & operation   &   & & forward path\\
                        
				 		\midrule
				 		{Ours} & \bf{10.97} & \bf{0.91} & \bf{22.9}5 & global & 2375-2375-250 & Full-Size & 1 \\
 					{Distort-and-Recover \cite{park2018distort}} &12.15&0.91& NA & global & 4750-250 & NA & multiple \\
 						\midrule
				 		
						
						{Ours} & {\bf{10.86}}& {\bf{0.91}} & {\bf{22.67}} & global & 2000-2000-1000 & Full-Size & 1 \\
					 					{Exposure \cite{hu2018exposure}} & 16.98& NA & NA & global & 2000-2000-1000 & Full-Size & multiple\\
 					 \midrule	
 					 	{Ours} & \bf{10.38} & \bf{0.93} & \bf{23.07} & global & 2250-2250-500 &  512 long side & 1 \\
						 					{Deep Photo Enhancer \cite{Chen:2018:DPE}{}} & NA& NA & 22.37& local & 2250-2250-500 & 512 long side & 1\\

						\bottomrule
					\end{tabular}
					\caption{
				A comparison of image enhancement methods trained with unpaired data. Results are evaluated by mean $L_2$, SSIM, and PSNR with respect to the expert target using the MIT-Adobe FiveK dataset.
				}
			\label{tab:exp_unsupervised}
			\end{center}
			\vspace{-4mm}
\end{table*}

\subsubsection{Ablation Study}
\label{sec:exp_ablation_study}
In order to evaluate our training techniques, as detailed in Section~\ref{sec:improving_unpaired_training}, we report the following setups: {\em Complete Method:} our full model as described; {\em No shared weights:} $G_X$ and $G_Y$ do not share their weights; {\em First phase only:} training with only the first phase (without the second cycle-consistency phase); {\em First phase only with Dropout:} training only the first phase, but adding dropout, similar to the second phase;  {\em Complete Method without Dropout:} performance of our full model but without adding dropout to the second phase; {\em Raw method: } a single phase training, without dropout and without weight sharing.

We evaluated the result on train-test split 2000-2000-1000 using $256\times256$ image resolution from MIT-Adobe FiveK dataset. Table~\ref{tab:ablation} demonstrates the effect of the individual components to the success of the training process. Sharing weights seem to have the largest effect, especially when training in two phases. The second phase of training contributes a relatively small improvement. 

\begin{table}
		\centering
					\begin{tabular}{lc}
						\toprule
                        
                        method & $\downarrow$mean $L_2$ \\
                        
				 		\midrule
				 		
						{Complete Method} & 10.43 \\
						
							 {No shared weights} & 17.58\\

 					{First phase only} &10.75\\
 					{First phase only with Dropout} &10.74\\
 					{Complete Method without Dropout} &10.76 \\
 					
                    {Raw method} & 15.32\\
						\bottomrule
					\end{tabular}
					\caption{
				Evaluation of our unpaired training techniques. The mean $L_2$ for different training procedures on the 1000 testing images from the MIT-Adobe FiveK dataset using $256\times256$ image resolution. Note that the results of Tab.~\ref{tab:exp_unsupervised} are given for the full resolution.}
                \label{tab:ablation}
\end{table}

\subsection{User study} 
\label{sec:user_study}
\begin{figure}
  \includegraphics[width=0.5\textwidth]{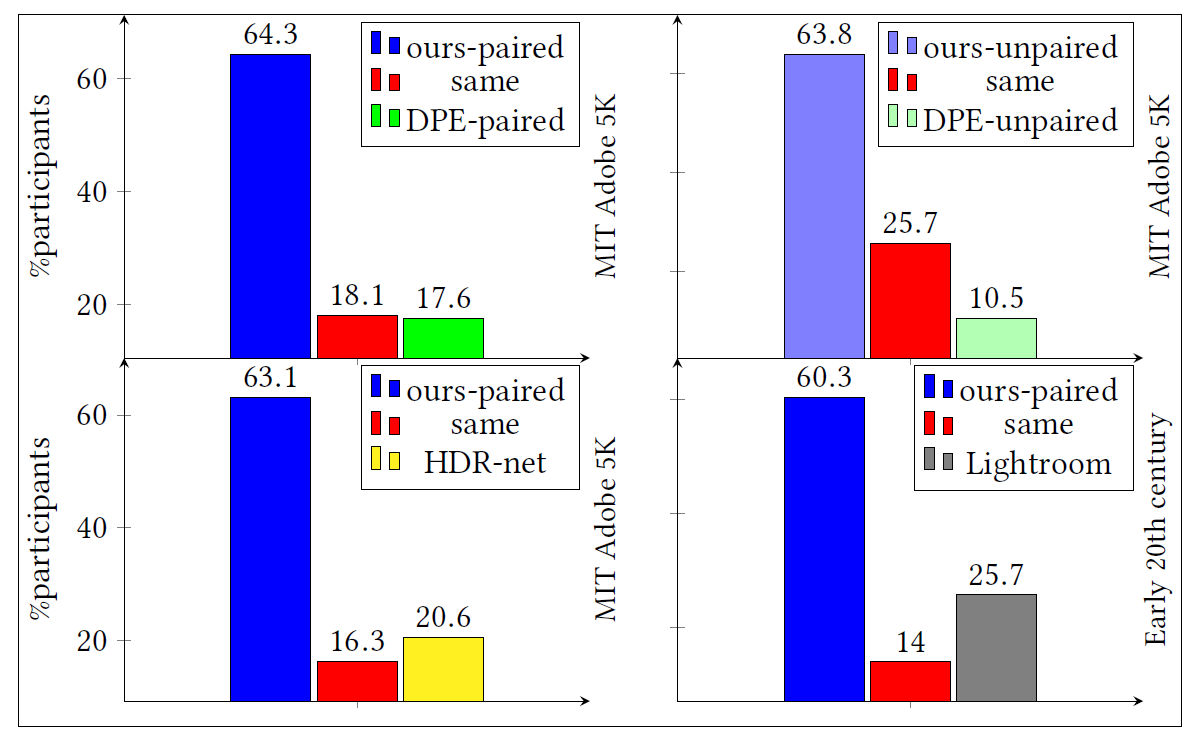}
  	\caption{
				{\em User study results.}
				Top: comparison to \cite{Chen:2018:DPE} in both the paired (left) and unpaired cases (right). Bottom-left: a comparison to \cite{gharbi2017deep} in the unpaired case. Bottom-right: a comparison to Adobe lightroom on old color photos. 
				Both~\cite{Chen:2018:DPE} and our methods were both trained with the same train-test split on MIT-ADOBE FiveK dataset. For comparing to~\cite{gharbi2017deep}, we used images from the test set of both methods.  }
  
  \label{fig:user_study}
\end{figure}

We conducted a user study to compare our retouching results to other leading methods with available code.
The user study was conducted on Amazon Mechanical Turk using pairwise comparisons and included 20 participants and 50 images. We presented to the users the images next to a zoom-in crop in a fixed location for each method. The users were asked to rate based on the visual colors and quality of the images, which image is better or if they are the same if they cannot decide. The images were randomly taken from the intersection of the method's test set. Figure~\ref{fig:user_study} describes the results of the user study. Both of our paired and unpaired models are clearly preferred by the users over the other tested methods.

\subsection{Applications}
Although our model was trained only on RAW input images, we  checked its generalization using some of the first color photos from the early 20th century\footnote{\href{https://www.boredpanda.com/first-color-photos-vintage-old-autochrome-lumiere-auguste-louis/?utm_source=google&utm_medium=organic&utm_campaign=organic}{https://www.boredpanda.com/first-color-photos-vintage-old-autochrome-lumiere-auguste-louis/}} and for a video scene from Game of Thrones, an American TV series.

\noindent{\bf Color photos from the early 20th:}
In order to evaluate our result, we randomly picked 50 images out of the available 116 images and compared them to the results of Adobe Lightroom, a leading commercial software. We used the Lightroom Auto-Tune feature to enhance the images. Samples of the enhancement results of both methods attached in supplementary materials.

We conducted a user study with 20 participants and 50 images using pairwise comparisons. We presented to the users the output of our method next to the Lightroom result. The users were asked to rate them based on the visual colors by selecting which image is better or stating that they are the same if they cannot decide. Samples of the enhancement results of both methods attached in supplementary materials. Figure~\ref{fig:user_study} shows our superiority through the user study result.

\noindent{\bf Game of Thrones:}
Game of Thrones recently released an episode that quickly became known as ``the dark episode'' since it consisted of dark and muddy scenes. To evaluate the ability of our method to perform a correction out of the box, we apply it to each frame individually, without any modification. For comparison, we also enhanced the video frames using Adobe Lightroom. 
Video results and sample images are attached as supplementary materials. As can be seen, our method outputs a natural looking frame, while Lightroom emphasizes the artifact in the coded video. 

\section{Conclusions}
We presented an automatic photo enhancer, which transforms input images to be with the characteristics of a given target set. Our method employed a parameterized color mapping and can be trained using either supervised or unsupervised learning. To improve performance in the unsupervised case, we introduced multiple training techniques such as weight sharing with individual Batch Normalization and a two-phase training process. In both types of supervision, the method achieves state-of-the-art results on the MIT-adobe FiveK dataset and shows superiority in a user study comparing it to literature methods as well as to a commercial software.

Applying our method, once trained, requires only one forward pass, unlike the reinforcement learning methods \cite{hu2018exposure,park2018distort}. Moreover, its transformation is determined from a low-resolution version of the input image, which leads to low computational demands. Yet, it can be applied to arbitrary resolutions, unlike \cite{Chen:2018:DPE} who require an input with a given resolution. Moreover, it employs a smooth global mapping function that prevents artifacts, which is advantageous over all the local modification methods \cite{Chen:2018:DPE, gharbi2017deep,yan2016automatic}, which report artifacts in their limitation. 

Our method deviates from the current trend of learning complex mappings between the input domain and the output domain, with an ever increasing capacity, which is balanced by regularization and skip connections. Instead, we apply a simple parametric transformation and couple it with a learned mechanism that provides the precise transformation parameters to every input image. 

\section*{Acknowledgements}
This project has received funding from the European Research Council (ERC) under the European Unions Horizon 2020 research and innovation program (grant ERC CoG 725974) and from ERC-StG grant no. 757497 (SPADE).


{\small
\bibliographystyle{ieee}
\bibliography{final_paper}

\begin{thebibliography}{10}\itemsep=-1pt

\bibitem{bianco2009image}
S.~Bianco, G.~Ciocca, F.~Marini, and R.~Schettini.
\newblock Image quality assessment by preprocessing and full reference model
  combination.
\newblock In {\em Image Quality and System Performance VI}, volume 7242, page
  72420O. International Society for Optics and Photonics, 2009.

\bibitem{bianco2019learning}
S.~Bianco, C.~Cusano, F.~Piccoli, and R.~Schettini.
\newblock Learning parametric functions for color image enhancement.
\newblock In {\em International Workshop on Computational Color Imaging}, pages
  209--220. Springer, 2019.

\bibitem{bychkovsky2011learning}
V.~Bychkovsky, S.~Paris, E.~Chan, and F.~Durand.
\newblock Learning photographic global tonal adjustment with a database of
  input/output image pairs.
\newblock In {\em CVPR 2011}, pages 97--104. IEEE, 2011.

\bibitem{Chen:2018:DPE}
Y.-S. Chen, Y.-C. Wang, M.-H. Kao, and Y.-Y. Chuang.
\newblock Deep photo enhancer: Unpaired learning for image enhancement from
  photographs with gans.
\newblock In {\em Proceedings of IEEE International Conference on Computer
  Vision and Pattern Recognition (CVPR 2018)}, pages 6306--6314, June 2018.

\bibitem{fairchild2013color}
M.~D. Fairchild.
\newblock {\em Color appearance models}.
\newblock John Wiley \& Sons, 2013.

\bibitem{faridul2014survey}
H.~S. Faridul, T.~Pouli, C.~Chamaret, J.~Stauder, A.~Tr{\'e}meau, E.~Reinhard,
  et~al.
\newblock A survey of color mapping and its applications.
\newblock {\em Eurographics (State of the Art Reports)}, 3, 2014.

\bibitem{gharbi2017deep}
M.~Gharbi, J.~Chen, J.~T. Barron, S.~W. Hasinoff, and F.~Durand.
\newblock Deep bilateral learning for real-time image enhancement.
\newblock {\em ACM Transactions on Graphics (TOG)}, 36(4):118, 2017.

\bibitem{hu2018exposure}
Y.~Hu, H.~He, C.~Xu, B.~Wang, and S.~Lin.
\newblock Exposure: A white-box photo post-processing framework.
\newblock {\em ACM Transactions on Graphics (TOG)}, 37(2):26, 2018.

\bibitem{hwang2012context}
S.~J. Hwang, A.~Kapoor, and S.~B. Kang.
\newblock Context-based automatic local image enhancement.
\newblock In {\em European Conference on Computer Vision}, pages 569--582.
  Springer, 2012.

\bibitem{ioffe2015batch}
S.~Ioffe and C.~Szegedy.
\newblock Batch normalization: Accelerating deep network training by reducing
  internal covariate shift.
\newblock {\em arXiv preprint arXiv:1502.03167}, 2015.

\bibitem{isola2017image}
P.~Isola, J.-Y. Zhu, T.~Zhou, and A.~A. Efros.
\newblock Image-to-image translation with conditional adversarial networks.
\newblock In {\em Proceedings of the IEEE conference on computer vision and
  pattern recognition}, pages 1125--1134, 2017.

\bibitem{kim2017learning}
T.~Kim, M.~Cha, H.~Kim, J.~K. Lee, and J.~Kim.
\newblock Learning to discover cross-domain relations with generative
  adversarial networks.
\newblock In {\em Proceedings of the 34th International Conference on Machine
  Learning-Volume 70}, pages 1857--1865. JMLR. org, 2017.

\bibitem{kingma2014adam}
D.~P. Kingma and J.~Ba.
\newblock Adam: A method for stochastic optimization.
\newblock {\em arXiv preprint arXiv:1412.6980}, 2014.

\bibitem{lee2016automatic}
J.-Y. Lee, K.~Sunkavalli, Z.~Lin, X.~Shen, and I.~So~Kweon.
\newblock Automatic content-aware color and tone stylization.
\newblock In {\em Proceedings of the IEEE Conference on Computer Vision and
  Pattern Recognition}, pages 2470--2478, 2016.

\bibitem{liu2014autostyle}
Y.~Liu, M.~Cohen, M.~Uyttendaele, and S.~Rusinkiewicz.
\newblock Autostyle: Automatic style transfer from image collections to users'
  images.
\newblock In {\em Computer Graphics Forum}, volume~33, pages 21--31. Wiley
  Online Library, 2014.

\bibitem{maas2013rectifier}
A.~L. Maas, A.~Y. Hannun, and A.~Y. Ng.
\newblock Rectifier nonlinearities improve neural network acoustic models.
\newblock In {\em Proc. icml}, volume~30, page~3, 2013.

\bibitem{park2018distort}
J.~Park, J.-Y. Lee, D.~Yoo, and I.~So~Kweon.
\newblock Distort-and-recover: Color enhancement using deep reinforcement
  learning.
\newblock In {\em Proceedings of the IEEE Conference on Computer Vision and
  Pattern Recognition}, pages 5928--5936, 2018.

\bibitem{reinhard2001color}
E.~Reinhard, M.~Adhikhmin, B.~Gooch, and P.~Shirley.
\newblock Color transfer between images.
\newblock {\em IEEE Computer graphics and applications}, 21(5):34--41, 2001.

\bibitem{schwartz2018deepisp}
E.~Schwartz, R.~Giryes, and A.~M. Bronstein.
\newblock Deepisp: Toward learning an end-to-end image processing pipeline.
\newblock {\em IEEE Transactions on Image Processing}, 28(2):912--923, 2018.

\bibitem{srivastava2014dropout}
N.~Srivastava, G.~Hinton, A.~Krizhevsky, I.~Sutskever, and R.~Salakhutdinov.
\newblock Dropout: a simple way to prevent neural networks from overfitting.
\newblock {\em The Journal of Machine Learning Research}, 15(1):1929--1958,
  2014.

\bibitem{susstrunk1999standard}
S.~S{\"u}sstrunk, R.~Buckley, and S.~Swen.
\newblock Standard rgb color spaces.
\newblock In {\em Color and Imaging Conference}, volume 1999, pages 127--134.
  Society for Imaging Science and Technology, 1999.

\bibitem{wang2004image}
Z.~Wang, A.~C. Bovik, H.~R. Sheikh, E.~P. Simoncelli, et~al.
\newblock Image quality assessment: from error visibility to structural
  similarity.
\newblock {\em IEEE transactions on image processing}, 13(4):600--612, 2004.

\bibitem{yan2016automatic}
Z.~Yan, H.~Zhang, B.~Wang, S.~Paris, and Y.~Yu.
\newblock Automatic photo adjustment using deep neural networks.
\newblock {\em ACM Transactions on Graphics (TOG)}, 35(2):11, 2016.

\bibitem{yi2017dualgan}
Z.~Yi, H.~Zhang, P.~Tan, and M.~Gong.
\newblock Dualgan: Unsupervised dual learning for image-to-image translation.
\newblock In {\em Proceedings of the IEEE international conference on computer
  vision}, pages 2849--2857, 2017.

\bibitem{CycleGAN2017}
J.-Y. Zhu, T.~Park, P.~Isola, and A.~A. Efros.
\newblock Unpaired image-to-image translation using cycle-consistent
  adversarial networkss.
\newblock In {\em Computer Vision (ICCV), 2017 IEEE International Conference
  on}, 2017.

\end{thebibliography}
}

\end{document}